\documentclass[sigconf,authorversion]{acmart}

\AtBeginDocument{%
  }

\acmConference[KDD-UC '24]{Make sure to enter the correct
  conference title from your rights confirmation emai}{August 25--29,
  2024}{Barcelona, Spain}
\acmISBN{978-1-4503-XXXX-X/18/06}

\usepackage{subfig}
\begin{document}



\title{Snowy Scenes,Clear Detections: A Robust Model for Traffic Light Detection in Adverse Weather Conditions}

\author{Shivank Garg}
\authornote{First three authors contributed equally to this research.}
\email{shivank_g@mfs.iitr.ac.in}
\authornotemark[1]
\affiliation{%
   \institution{Indian Institute of Technology Roorkee}
   \city{Roorkee}
   \country{India}
 }

 \author{Abhishek Baghel}
 \email{abhishek_b@mfs.iitr.ac.in}
 \affiliation{%
   \institution{Indian Institute of Technology Roorkee}
   \city{Roorkee}
   \country{India}
 }

\author{Amit Agarwal}
\email{aagarwal3@cs.iitr.ac.in}
\affiliation{%
   \institution{Indian Institute of Technology Roorkee.}
   \city{Roorkee}
   \country{India}
 }

\author{Durga Toshniwal}
 \email{durga.toshniwal@cs.iitr.ac.in}
 \affiliation{%
  \institution{Indian Institute of Technology Roorkee}
  \city{Roorkee}
  \country{India}}


\begin{abstract}

With the rise of autonomous vehicles and advanced driver-assistance systems (ADAS), ensuring reliable object detection in all weather conditions is crucial for safety and efficiency. Adverse weather like snow, rain, and fog presents major challenges for current detection systems, often resulting in failures and potential safety risks. This paper introduces a novel framework and pipeline designed to improve object detection under such conditions, focusing on traffic signal detection where traditional methods often fail due to domain shifts caused by adverse weather. We provide a comprehensive analysis of the limitations of existing techniques. Our proposed pipeline significantly enhances detection accuracy in snow, rain, and fog. Results show a 40.8\% improvement in average IoU and F1 scores compared to naive fine-tuning and a 22.4\% performance increase in domain shift scenarios, such as training on artificial snow and testing on rain images.
\end{abstract}






\maketitle
\begin{figure*}
  \includegraphics[width=\textwidth]{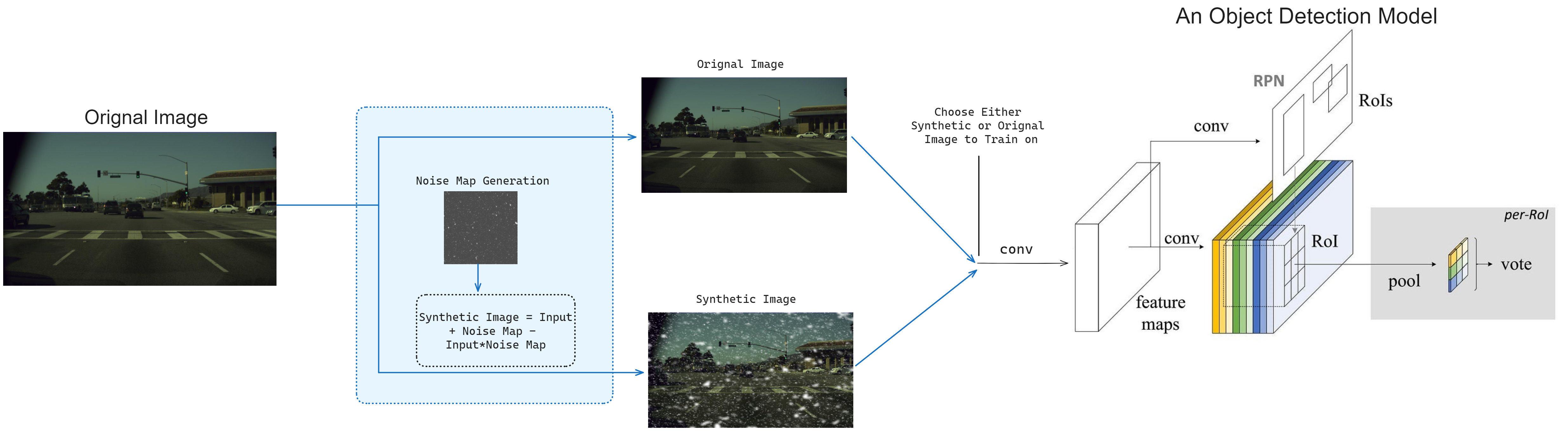}
  \caption{Our Pipeline: We create a synthetic version of the ground truth image and use either the original or the synthesized one for training the model.}
  \Description{We first create synthetic images from real images and then we further randomly sample and choose either of the images to fine-tune our object detection model}
  \label{fig:approach}
\end{figure*}
\section{Introduction}

In recent years, the rise of autonomous vehicles has significantly increased the demand for reliable object detection systems \cite{zaidi2021survey} that can operate effectively under various weather conditions. Ensuring the accurate identification of traffic signals during adverse weather, such as snow, is essential for the safe operation of self-driving cars. Maintaining high detection accuracy in these challenging conditions can greatly enhance the safety and reliability of autonomous driving systems \cite{badue2021self}, helping to prevent accidents and improve overall traffic safety. Beyond autonomous vehicles, robust object detection models are crucial for other critical applications, including advanced driver-assistance systems (ADAS), intelligent traffic management, and disaster response robotics, where precise and reliable visual recognition under varying environmental conditions is vital.

The importance of dependable traffic signal detection is highlighted by real-world incidents. For instance, during testing periods in California, autonomous vehicles have experienced accidents due to the failure to correctly detect traffic signals. In some cases, these failures have led to disengagements where human drivers had to take control to avoid potential collisions\cite{favaro2017examining}. Reports indicate that Google's autonomous vehicles recorded several disengagements due to improper perception of traffic lights, which could have resulted in accidents if not promptly managed by safety drivers\cite{dixit2016autonomous}. Moreover, a comprehensive survey \cite{van2018autonomous} on vehicle perception underscores the ongoing need to improve vehicular perception, particularly for object detection in poor weather conditions, and suggests that data fusion could address this issue.

Over the past decade, Convolutional Neural Networks (CNNs) \cite{li2021survey} have become essential tools for tasks such as object detection\cite{zaidi2021survey}, image segmentation \cite{cheng2001color}, and image restoration \cite{banham1997digital}. Despite their widespread use, these deep learning models are largely data-driven and tend to overfit their training data \cite{rice2020overfitting}, which can result in significant performance drops when faced with datasets that differ substantially from the training data. This issue is particularly crucial in high-stakes applications like autonomous driving, where dependable object detection is vital to prevent accidents and ensure safety. The growing importance of robust object detection in such sensitive tasks underscores the need to address the challenges posed by differing data distributions.

This research introduces a novel pipeline and framework aimed at improving model training by incorporating both synthetic and real images. This method seeks to enhance overall accuracy, especially when handling test data that significantly diverges from the training distribution. The proposed approach demonstrates potential for increased reliability and safety in critical real-world applications. Notably, this work is among the first to address the detection of traffic signals in snowy conditions.

A new traffic light dataset has been created, comprising both synthetic and real images. The real images are based on the Bosch Small Traffic Light Dataset \cite{BehrendtNovak2017ICRA} , while the synthetic images are generated using the approach by \cite{zhang2023weatherstream}.

Several state-of-the-art object detection models, including Detectron 2 \cite{wu2019detectron2}, YOLO v7 \cite{wang2023yolov7}, and YOLO v8 \cite{Jocher_Ultralytics_YOLO_2023}, are evaluated using this dataset. Performance comparisons between the ground truth dataset and the artificially generated images indicate that the highest detection accuracy is achieved with the newly created dataset. Additionally, experiments demonstrate that models trained using this pipeline also perform well in real-life environments with fire, fog, smog, and snow \cite{zhang2021adaptive}, suggesting robust applicability across various adverse weather conditions.

The proposed pipeline is versatile, accommodating various noise types and image editing techniques, such as those based on Generative Adversarial Networks (GANs) \cite{goodfellow2014generative}, diffusion models \cite{ho2020denoising}, and OpenCV \cite{bradski2000opencv}. This flexibility allows for the synthesis of adverse weather images and the application of different models for fine-tuning. While this study focuses on traffic light detection under hazing conditions, the pipeline can be extended to other tasks as well. The code for our work is available at \href{https://github.com/abhishekbaghel11/Traffic-light-detection-in-adverse-weather-conditions.git}{\textcolor{magenta}{Link to code}}
\vspace{-5pt}
\section{Related Works}
Recent advancements in deep learning have revolutionized the field of image editing \cite{bayar2016deep}, enabling sophisticated modifications with high levels of realism and precision. Techniques such as Generative Adversarial Networks (GANs)\cite{goodfellow2014generative} and diffusion models\cite{ho2020denoising} have proven particularly effective in generating and altering images . These methods can introduce artificial elements, remove noise, and enhance image quality, making them invaluable for applications in various domains, including improving the robustness for traffic signal detection under adverse weather conditions.

\subsection{Object Detection during Bad weather condition: }

    There are several examples of deep learning-based identification tasks for autonomous driving-related objects like cars, people, traffic lights, drivable pathways, or lanes. For instance, deep learning frameworks were utilized by \cite{tao2022vehicle} to detect vehicles in foggy conditions using an attention module for better feature extraction. The ZUT dataset offered by \cite{tumas2020pedestrian} and the YOLOv3\cite{redmon2018yolov3} technique were employed to identify pedestrians in adverse weather conditions, including rain, fog, and frost. Further, YOLO approaches have been used to detect pedestrians in hazy weather\cite{li2019deep}. More examples of deep learning networks for object detection in adverse weather conditions include the dual subnet network (DSNet)\cite{huang2020dsnet} that jointly learned visibility enhancement, object classification, and localization \cite{qin2022denet}, and DENet, an adaptive image enhancement model trained with YOLOv3 for better detection in challenging conditions.

\subsection{Recent Technique for object detection:}
\subsubsection{Faster RCNN}
Faster R-CNN \cite{ren2015faster} \cite{girshick2015fast}is a object detection model that integrates deep convolutional networks with Region Proposal Networks (RPNs) to achieve high-speed and accurate object detection. The model's architecture effectively combines feature extraction and region proposal into a single, unified framework, enhancing both efficiency and performance . Its design set a new object detection standard, demonstrating superior capabilities in various applications.

\subsubsection{YOLO}
The YOLO ("You Only Look Once") \cite{redmon2016you} model processes an input image using a deep convolutional neural network to detect objects within the image. Initially, the model's convolutional layers are pre-trained on the ImageNet dataset by incorporating a temporary average pooling layer and a fully connected layer. Following this pre-training phase, the model is adapted and fine-tuned specifically for object detection tasks.

\subsubsection{YOLO V7}
The YOLO v7 \cite{wang2023yolov7}model significantly outperforms previous YOLO-based models. Unlike its predecessors, YOLO v7 employs a focal loss function rather than the traditional cross-entropy loss, enhancing its detection capabilities. Additionally, it utilizes a set of nine predefined anchor boxes for more efficient image detection and operates at a higher resolution, further improving its accuracy and performance.

\subsubsection{YOLO V8}
YOLO v8  \cite{Jocher_Ultralytics_YOLO_2023}builds upon the strengths of its predecessors, enhancing both speed and accuracy in object detection. It utilizes the CSPDarknet53 architecture as its backbone and incorporates a Path Aggregation Network to integrate information from different scales within the image. Additionally, YOLO v8 introduces more efficient feature extraction methods and improved anchor-free detection mechanisms, enabling superior performance with reduced computational resource requirements.

\section{Method and Experiments}
We propose an innovative pipeline to fine-tune and train object detection models, achieving higher accuracy even when there is a high difference between the actual test data and the training data. Our work in particular, focuses on the challenging task of detecting traffic lights in adverse, snowy weather conditions. Our approach involves using ground truth images and labels to generate synthetic images [Section 3.2]. These synthetic images are then used to fine-tune our models, with detailed parameters and model choices explained in [Section 3.3]. The structure of our pipeline is illustrated in [Figure \ref{fig:approach}].

\begin{figure*}[!t]
    \centering
    \subfloat[Snowfall(Baseline)]{\includegraphics[width=2.5cm,height=2.5cm]{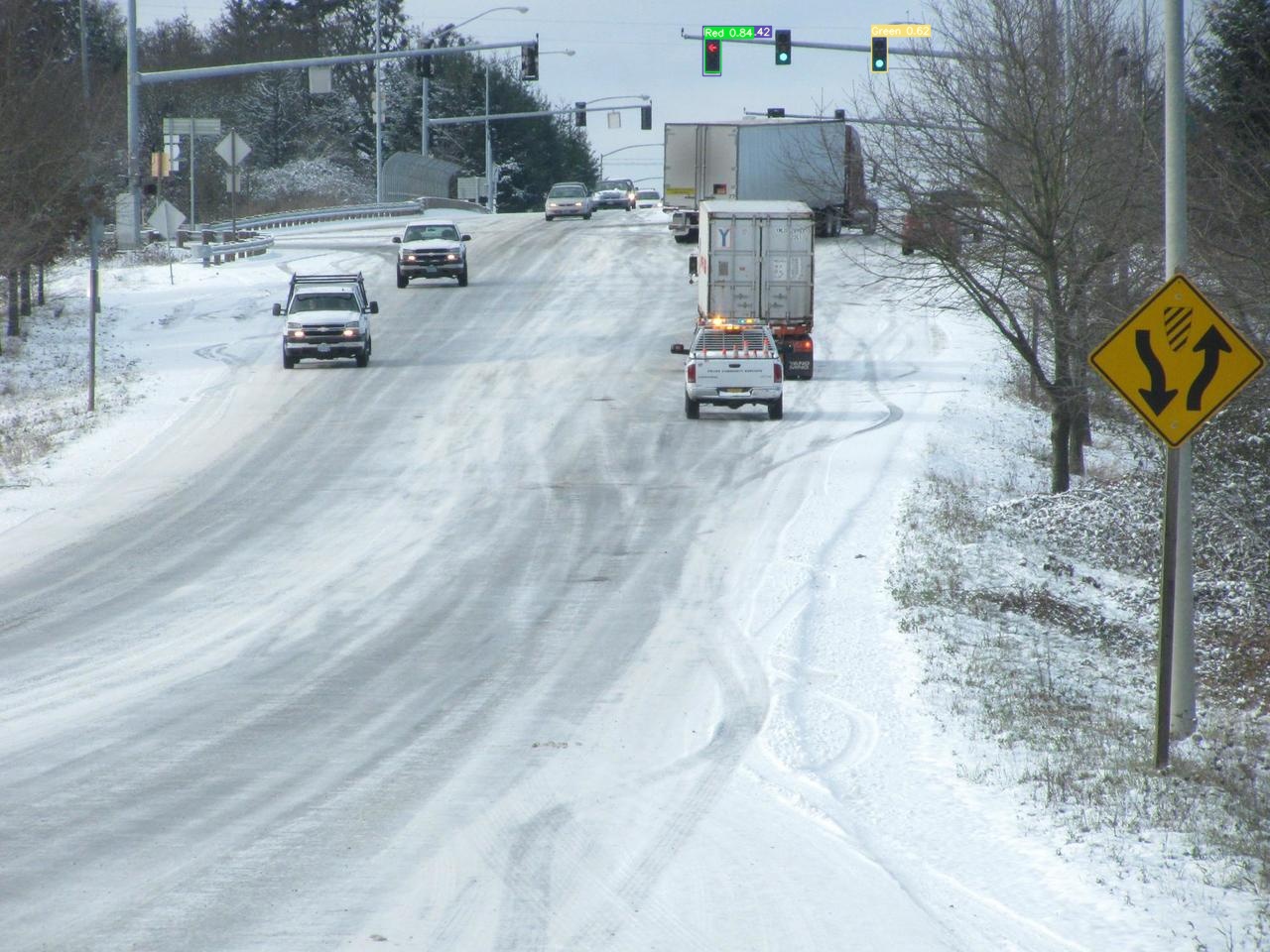}}%
    \label{f1}
    \hspace{1mm}
    \subfloat[Snowfall(Ours)]{\includegraphics[width=2.5cm,height=2.5cm]{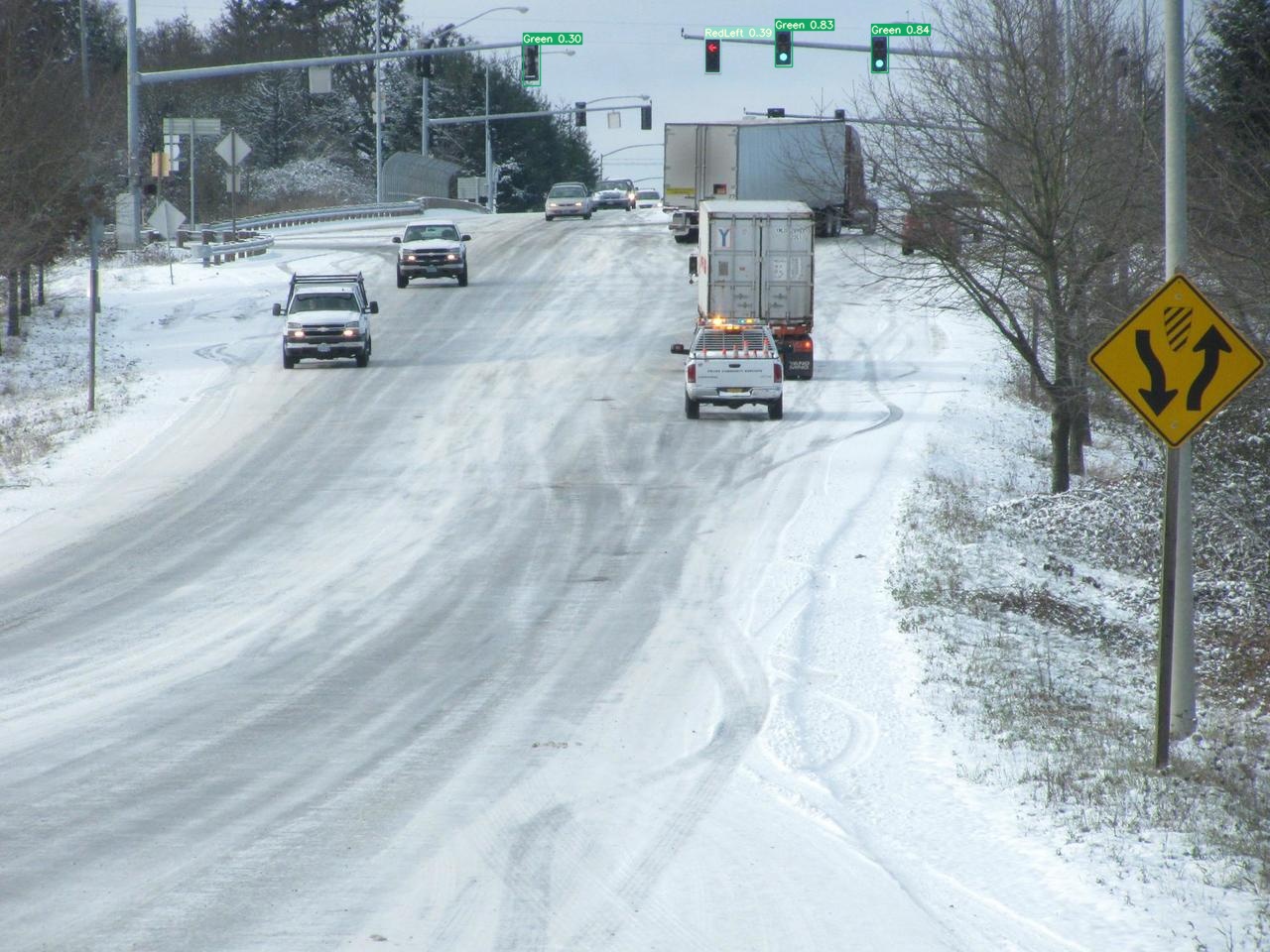}}%
    \label{f2}
    \hspace{5mm} 
    \subfloat[Snowfall(Baseline)]{\includegraphics[width=2.5cm,height=2.5cm]{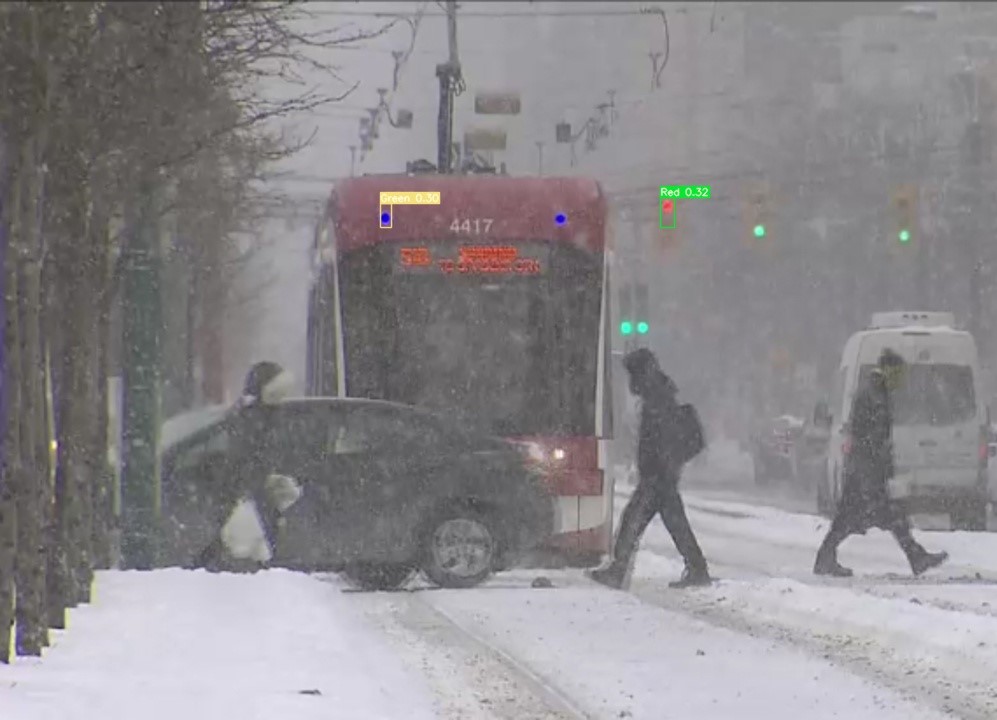}}%
    \label{f3}
    \hspace{1mm}
    \subfloat[Snowfall(Ours)]{\includegraphics[width=2.5cm,height=2.5cm]{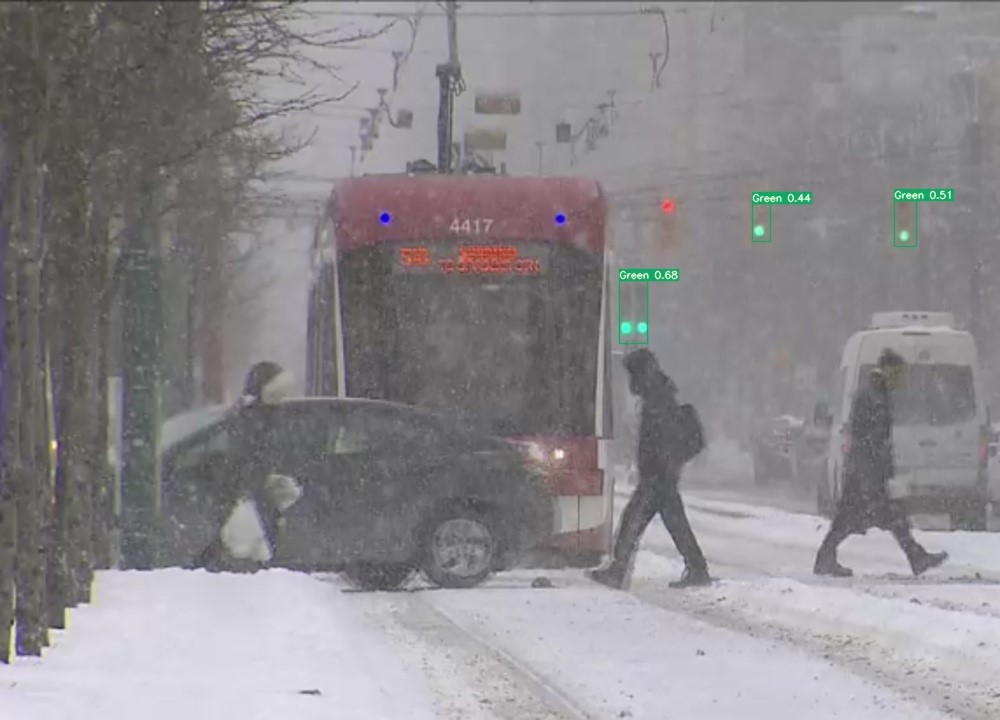}}%
    \label{f4}
    \hspace{5mm}
    \subfloat[Fog(Baseline)]{\includegraphics[width=2.5cm,height=2.5cm]{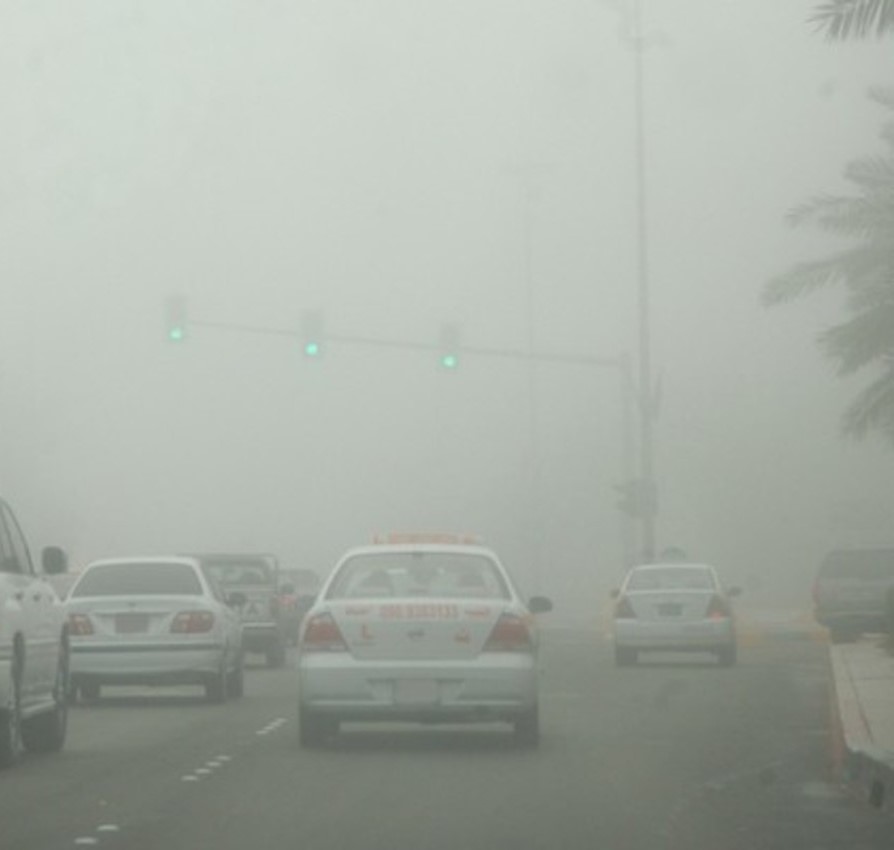}}%
    \label{f5}
    \hspace{1mm}
    \subfloat[Fog(Ours)]{\includegraphics[width=2.5cm,height=2.5cm]{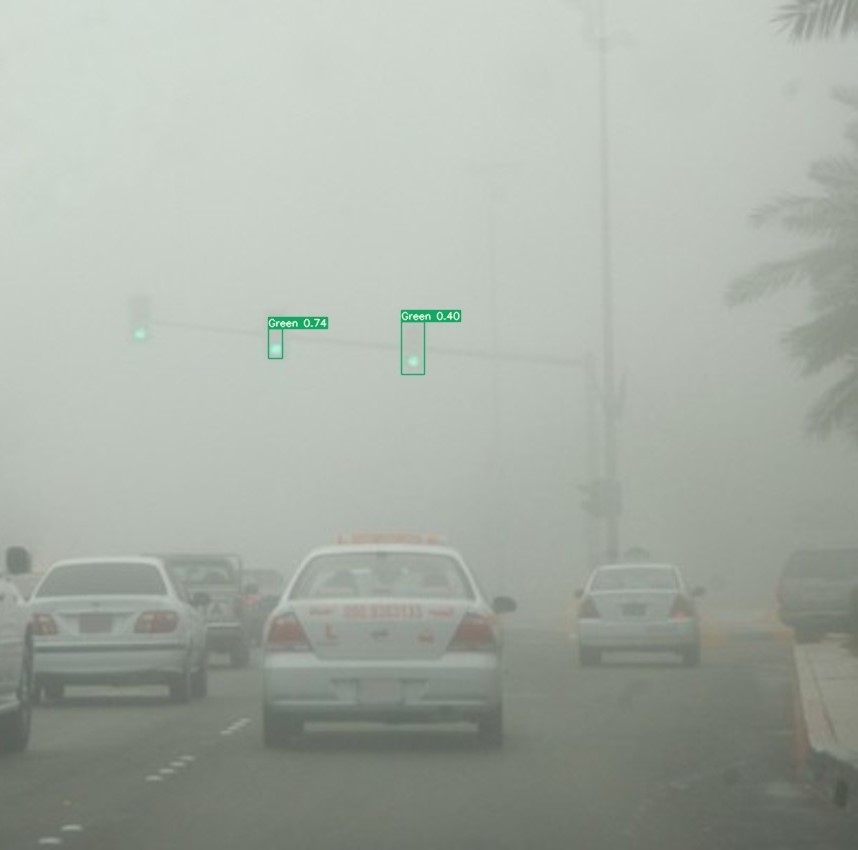}}%
    \label{f6}    
    \\
    \subfloat[Smog(Baseline)]{\includegraphics[width=2.5cm,height=2.5cm]{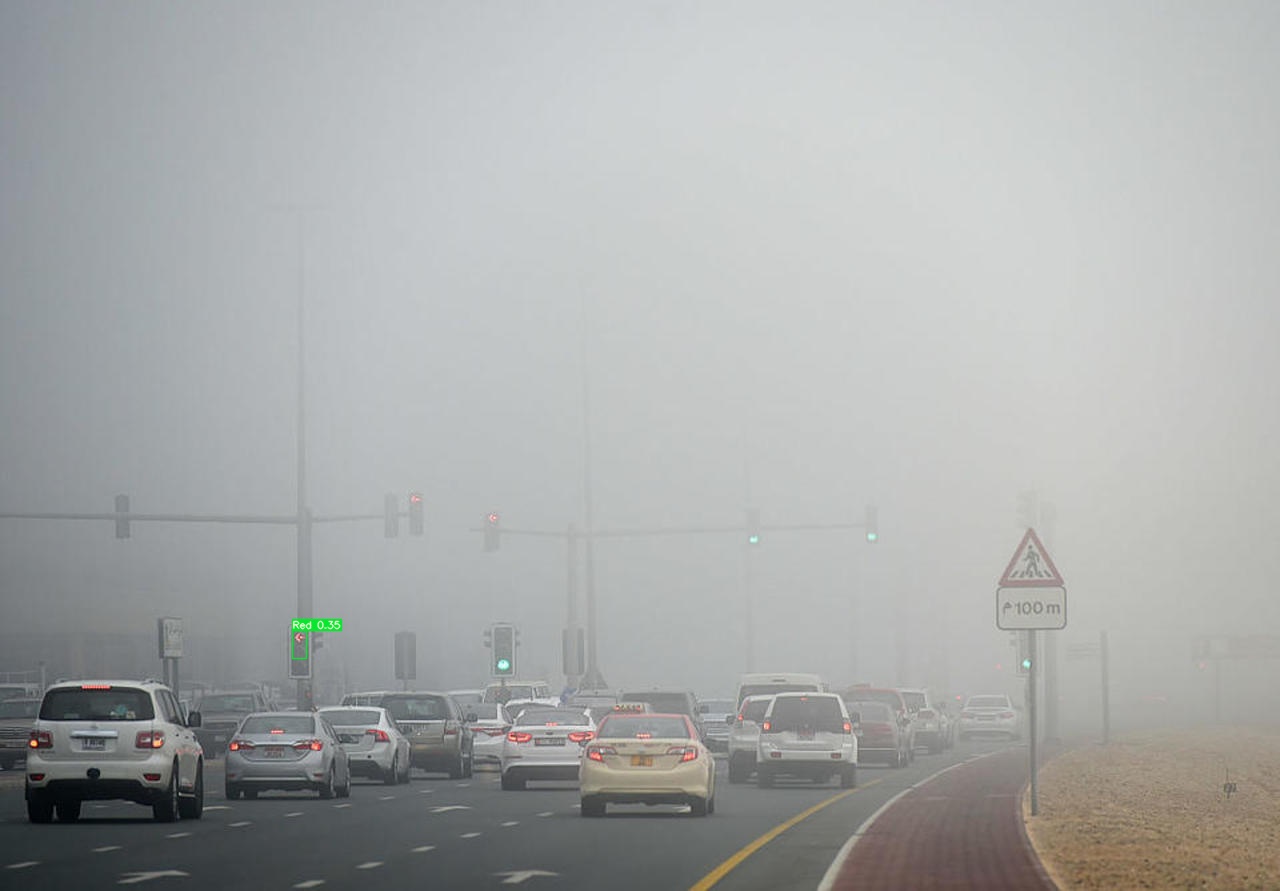}}%
    \label{f7}
    \hspace{1mm}
    \subfloat[Smog(Ours)]{\includegraphics[width=2.5cm,height=2.5cm]{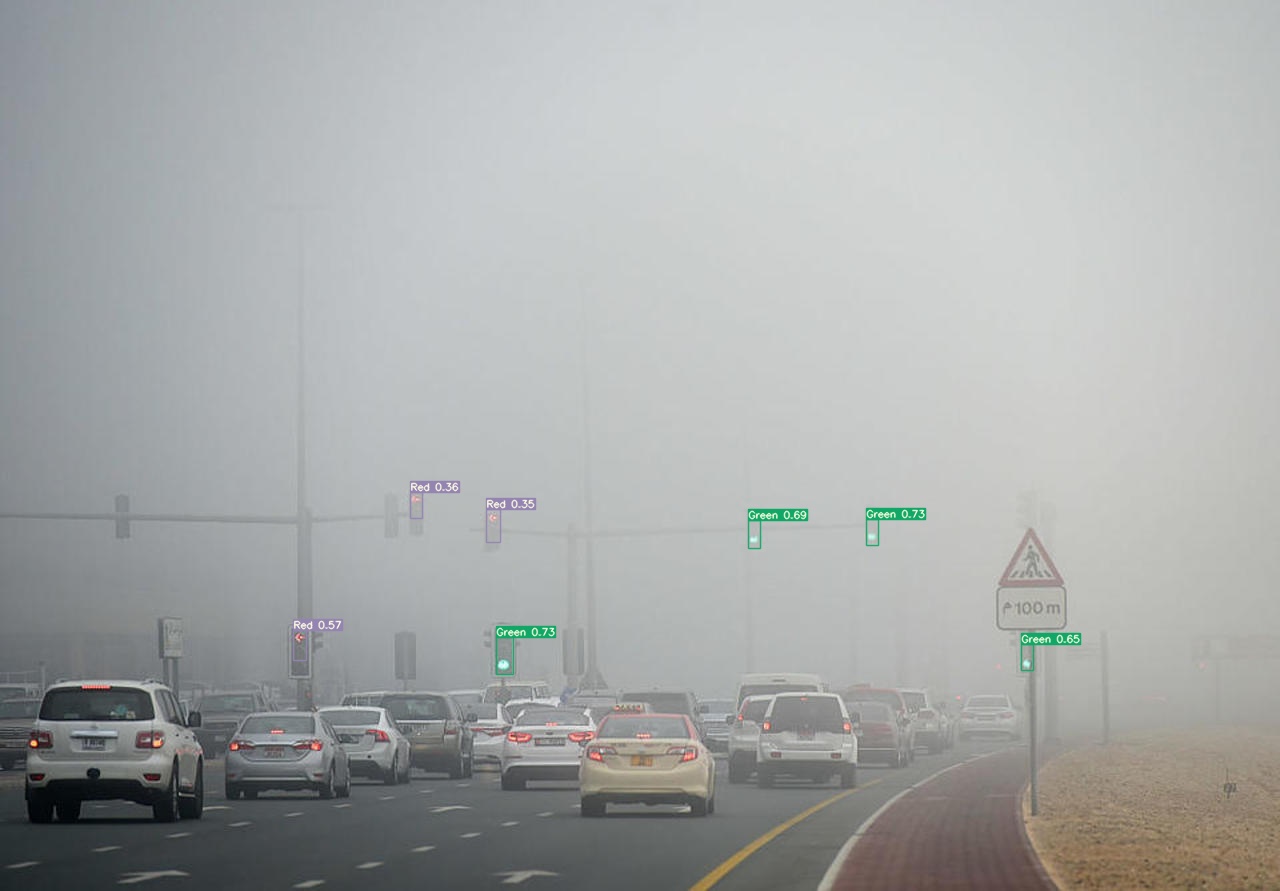}}%
    \label{f8}
    \hspace{5mm}
    \subfloat[Fire(Baseline)]{\includegraphics[width=2.5cm,height=2.5cm]{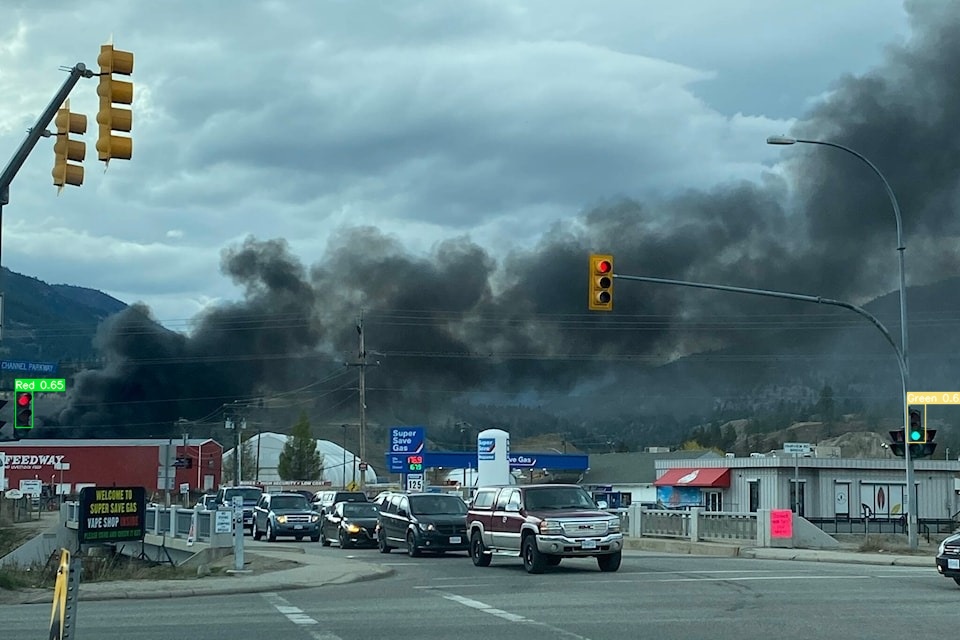}}%
    \label{f9}
    \hspace{1mm}
    \subfloat[Fire(Ours)]{\includegraphics[width=2.5cm,height=2.5cm]{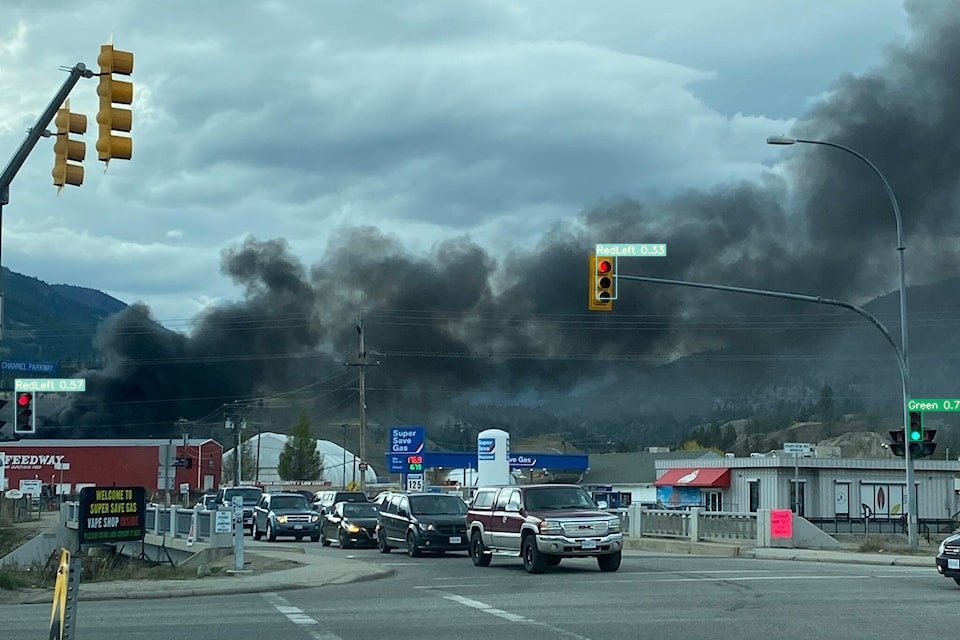}}%
    \label{f10}
    \hspace{5mm}
    \subfloat[Rainfall(Baseline)]{\includegraphics[width=2.5cm,height=2.5cm]{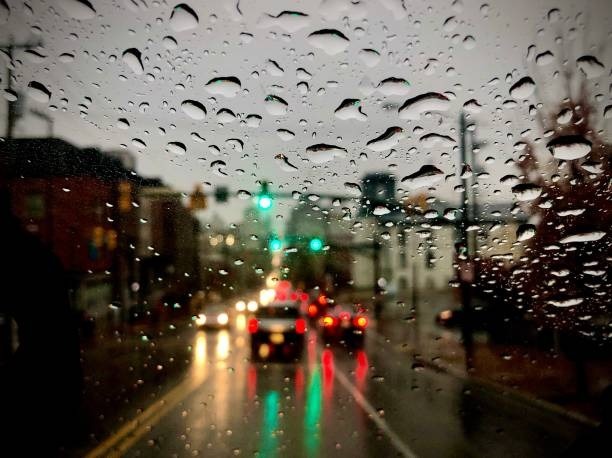}}%
    \label{f11}
    \hspace{1mm}
    \subfloat[Rainfall(Ours)]{\includegraphics[width=2.5cm,height=2.5cm]{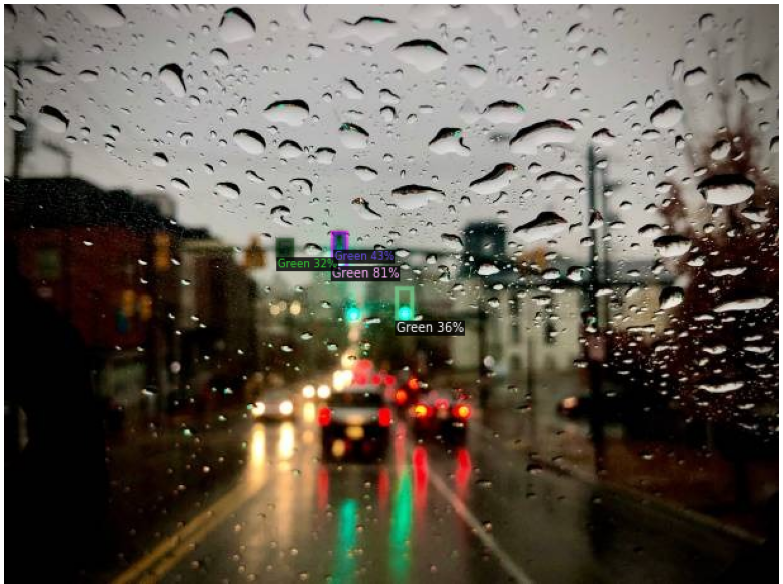}}%
    \label{f12}
    \caption{Comparison of YOLO v8 tested for naturally occurring real images. Left: Images tested on model fine-tuned on the proposed pipeline for artificial snow dataset. Right: Images tested on model model fine-tuned on ground truth traffic light dataset}
    \label{fig:informative}
\end{figure*}

\subsection{Dataset}
To train and test our model, we use the Bosch Small Traffic Light Dataset \cite{BehrendtNovak2017ICRA}, which consists of 13,427 camera images at a resolution of 1280x720 pixels, and approximately 24,000 annotated traffic lights. For our experiments, we consider a subset of 1,600 images to train our models. 

\subsection{Synthetic Dataset Generation}
Based on the approach detailed in the \cite{zhang2023weatherstream} paper, we developed a method to create a synthetic dataset of snowy images. The approach leverages principles from light transport to automate the generation of high-quality image pairs, ensuring diverse and realistic weather conditions. The method utilizes light transport principles to effectively simulate and curate weather conditions, particularly snow, without the need for manual labeling which was required by previous works. The process involves the following key steps:

\begin{enumerate}
\item \textbf{Resize the Image and Initialize variables}: Resize the image to a fixed shape. Select the mean and standard deviation for sampling from the Gaussian distribution. Determine the number of different particle sizes to add to the image using the values in SCALE\_ARRAY (we use five different scales). Iterate over the different scales and during each iteration, perform the following steps:
\item \textbf{Create Noise Array}: Create a noise array from the selected mean and standard deviation using a Gaussian distribution.
\item \textbf{Apply Gaussian Filter}: Smooth the noise array using a Gaussian filter \cite{young1995recursive}. Apply a threshold value, setting all values above it to one and all values below it to zero.
\begin{equation}
G(x, y) = \frac{1}{2\pi\sigma^2} \exp \left( -\frac{x^2 + y^2}{2\sigma^2} \right)
\end{equation}
Where $\sigma$ is the standard deviation of the Gaussian distribution, and x and y are the spatial coordinates.
\item \textbf{Motion Blur}: Apply a motion blur \cite{gong2017motion} to the noise array using a kernel. The kernel is created by first generating an array with a horizontal line and then rotating it by a random angle to apply the motion blur in random directions. The kernel is smoothed according to the scales using a Gaussian filter. If the scale is small, the kernel is smoothed more; if the scale is large, the smoothing is less; and for moderate scales, the kernel is moderately smoothed.
\item \textbf{Blend Noise Layer}: Blend the obtained noisy layer with the input image for each iteration using the formula:
\begin{equation}
{input\_img}^k = {input\_img}^{k-1} \times (1 - {layer}) + {layer} \times 255  
\end{equation}
\item \textbf{Resize to Original Dimensions}: Resize the final image back to the original dimensions and return it.
\end{enumerate}
    
    
    
    
    

\subsection{Fine Tuning}

To demonstrate the efficiency of our proposed pipeline, we conducted extensive experiments evaluating the performance of Faster-RCNN pre-trained on COCO Dataset using Detectron 2, YOLO v7, and YOLO v8. We fine-tuned these pre-trained models on three different datasets: the ground truth dataset, the synthetic dataset, and the dataset generated using our proposed pipeline. The configurations for fine-tuning the models are detailed in [Table \ref{tab:configurations}].

\begin{table}[h]
    \centering
    \begin{tabular}{lccc}
        \hline
        & \textbf{YOLO v8} & \textbf{YOLO v7} & \textbf{Faster RCNN} \\
        \hline
        \textbf{Epochs} & 50 & 50 & 12000(Iterations) \\
        \textbf{Batch Size} & 16 & 16 & 2 \\
        \textbf{Learning Rate} & 0.01-0.1 & 0.01-0.1 & 0.001 \\
        \textbf{Scheduler} & Linear & Linear & None \\
        \hline
    \end{tabular}
    \caption{Configuration details for model fine-tuning}
    \label{tab:configurations}
\end{table}

In our experiments, we first fine-tuned each model on the ground truth dataset to establish a baseline performance. Next, we trained the models on the synthetic dataset, which was created using the method described earlier. Finally, we applied our proposed pipeline to generate an enhanced training dataset that includes both synthetic and real images with adverse weather conditions.

Our results indicate that the models achieve significantly better performance on the test data and real-life images when trained using our proposed pipeline. 

\section{Results}

\subsection{Validation Results}

In this work, all fine-tuned models were rigorously evaluated using a validation dataset that comprised a diverse mix of real, synthetic, and real snow images. The detailed results of these evaluations are presented in [Table \ref{tab:performance_comparison}]. Our evaluation metrics included Average Intersection over Union (IoU), mean Average Precision (mAP50-95) — which represents the mean mAP over IoU thresholds ranging from 0.5 to 0.95, map50 - which represents mAP over IoU threshold of 0.5, Precision (P) and F1 scores \cite{padilla2020survey}. These metrics were assessed on a substantial validation set containing 600 images. 

\begin{figure}[!t]
    \centering
    \subfloat[Ground Truth]{\includegraphics[width=2.5cm,height=2.5cm]{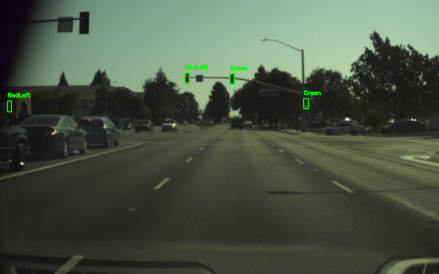}}%
        \label{fv1}
    \subfloat[Baseline]{\includegraphics[width=2.5cm,height=2.5cm]{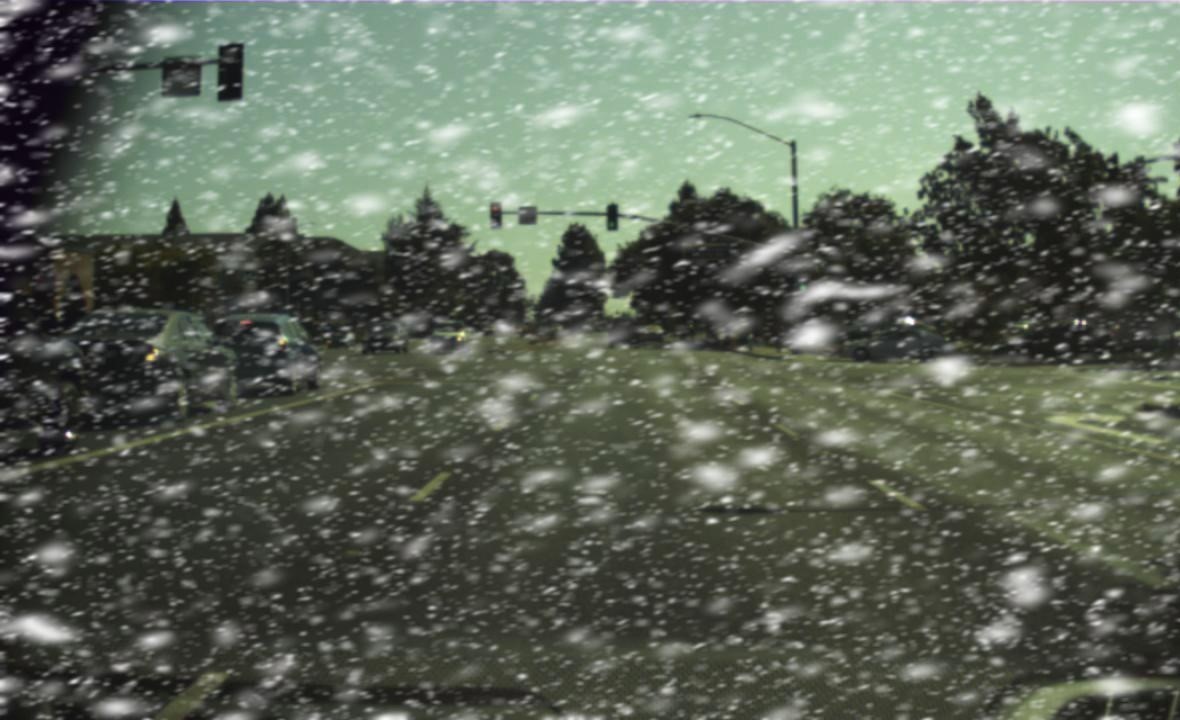}}%
        \label{fv2}
    \hspace{1mm}
    \subfloat[Ours]{\includegraphics[width=2.5cm,height=2.5cm]{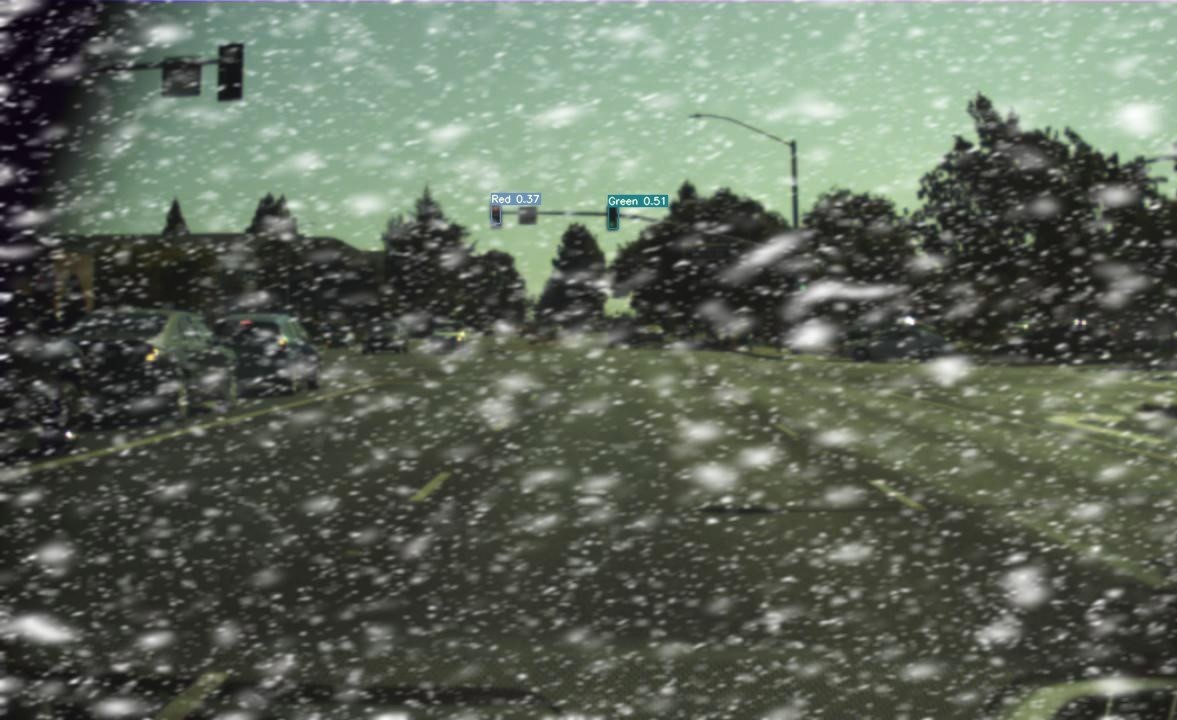}}%
        \label{fv3}
    \hspace{1mm}
    \\
    \subfloat[Ground Truth]{\includegraphics[width=2.5cm,height=2.5cm]{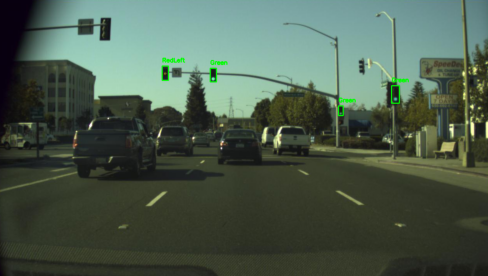}}%
        \label{fv4}
    \hspace{1mm}
    \subfloat[Baseline]{\includegraphics[width=2.5cm,height=2.5cm]{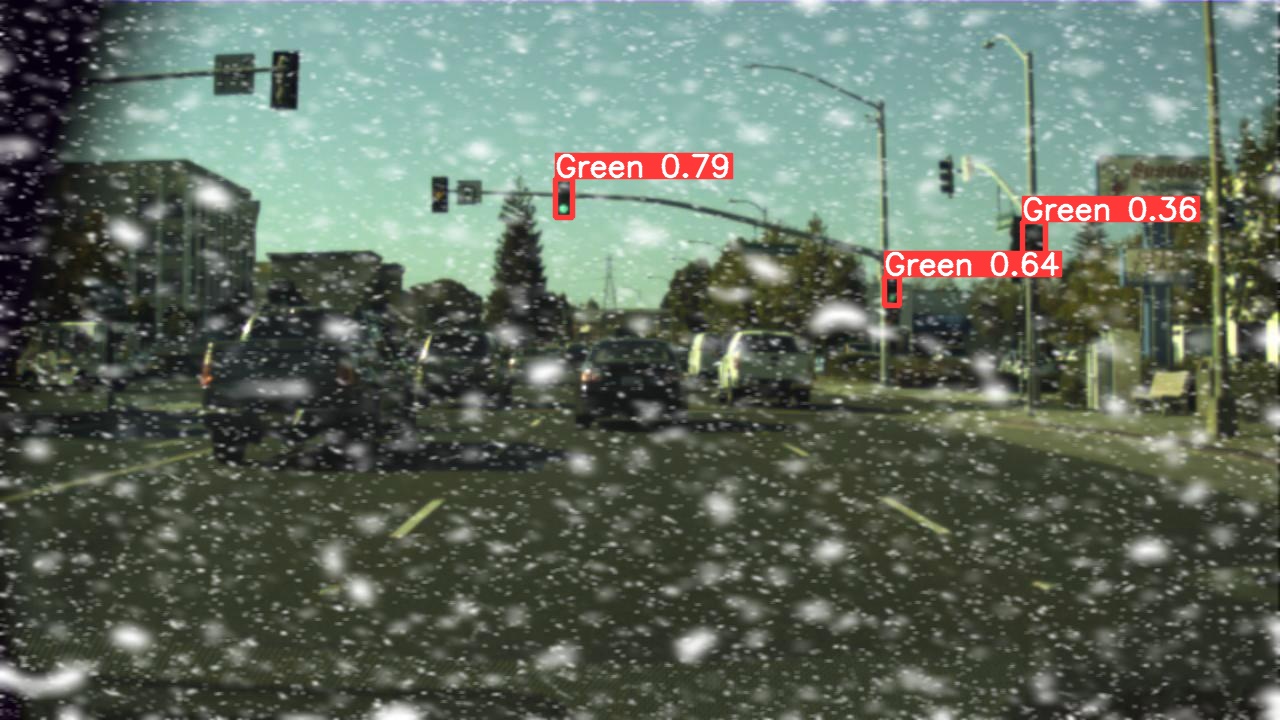}}%
        \label{fv5}
    \hspace{1mm}
    \subfloat[Ours]{\includegraphics[width=2.5cm,height=2.5cm]{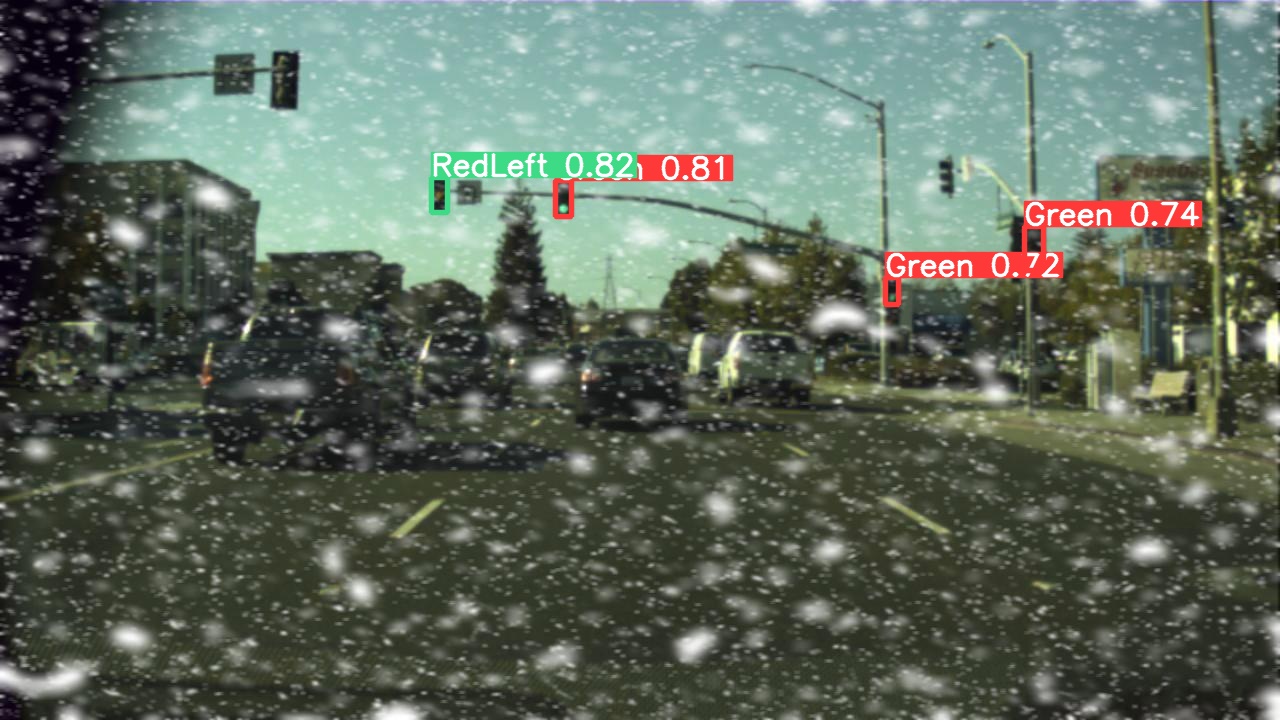}}%
        \label{fv6}
    \\
        \subfloat[Ground Truth]{\includegraphics[width=2.5cm,height=2.5cm]{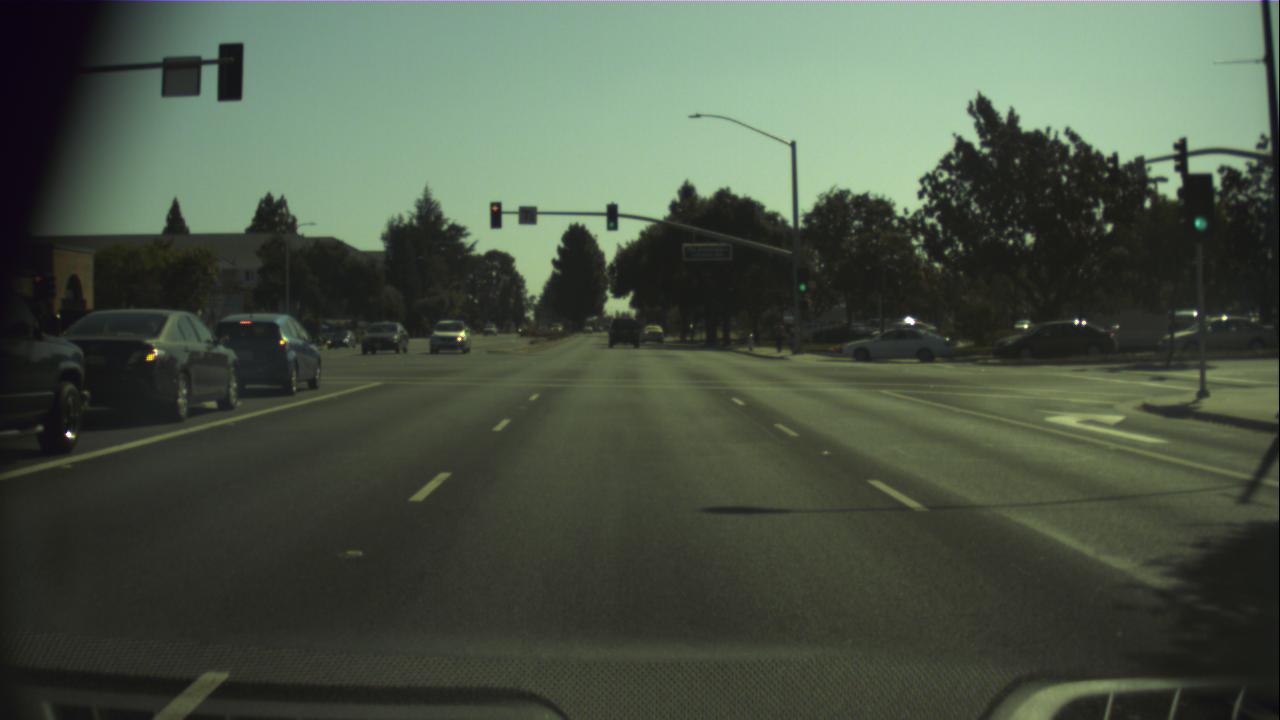}}%
        \label{fv7}
    \hspace{1mm}
    \subfloat[Baseline]{\includegraphics[width=2.5cm,height=2.5cm]{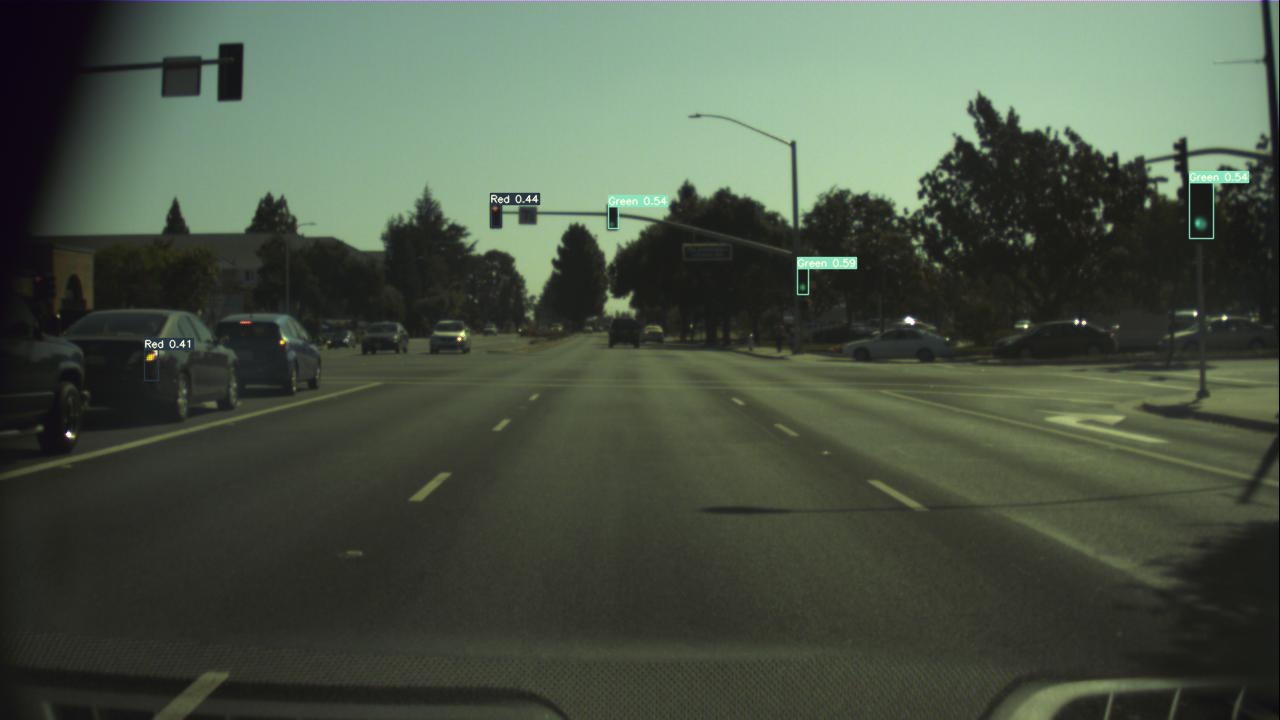}}%
        \label{fv8}
    \hspace{1mm}
    \subfloat[Ours]{\includegraphics[width=2.5cm,height=2.5cm]{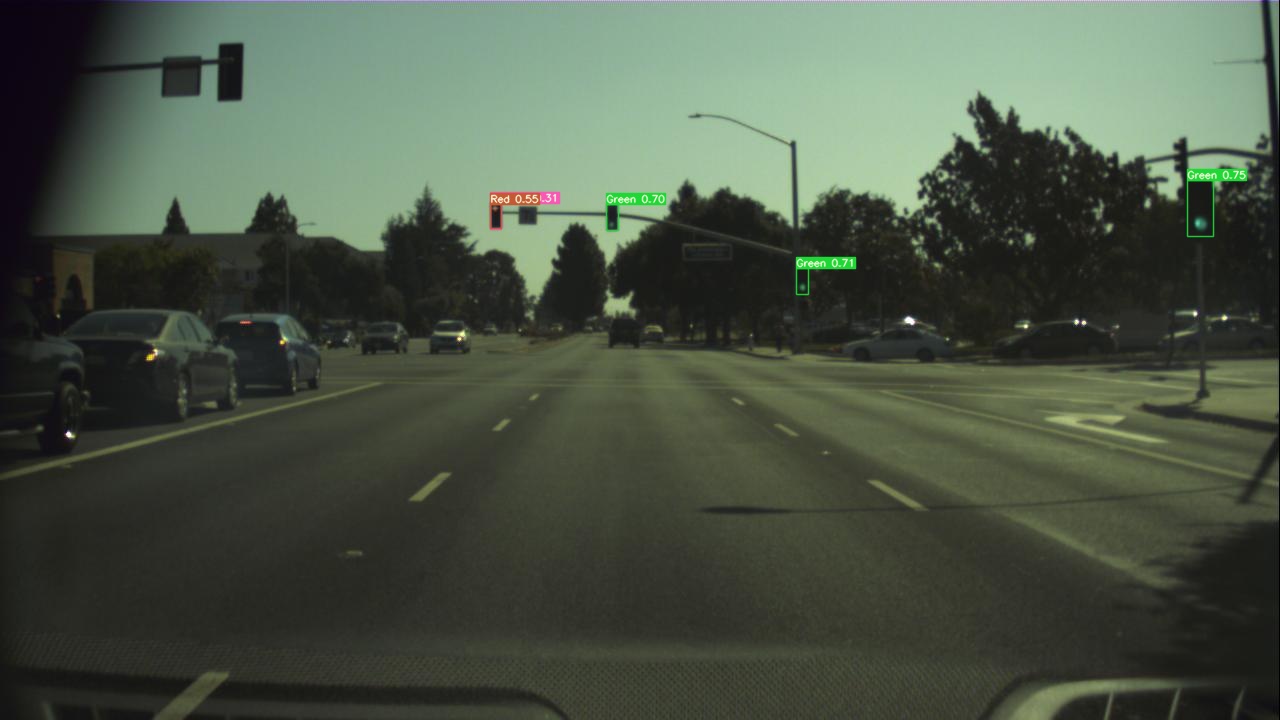}}%
        \label{fv9}
    \caption{Visual Comparison of Results on Validation Set Obtained by YOLO v8 Model}
    \label{fig:Validation}
\end{figure}

Based on our evaluations, the baseline model and the model fine-tuned using our pipeline performed similarly for images with light snowfall. However, for images with heavy snowfall, our models demonstrated significantly better performance compared to the baseline. Additionally, in scenarios without any snowfall, the baseline model occasionally misclassified car lights as traffic lights. In contrast, our model did not make such errors, likely due to the augmentation added during training when fine-tuning with our proposed pipeline. Some images from the validation set are shown in [Figure \ref{fig:Validation}].

The calculation of IoU was performed using the following methodology:

1. \textbf{IoU Calculation Between Two Boxes:} The IoU between two boxes is determined by the ratio of the area of intersection to the area of union of the two boxes.
   
2. \textbf{Image-Level IoU Calculation:} For an entire image, we calculate the IoU by first finding the IoU for each matched pair of ground truth and predicted boxes. A match is identified as the pair with the highest IoU, provided this IoU exceeds a threshold of 0.5. Each ground truth box is allowed to match with only one predicted box.

3. \textbf{Summing IoU of Matches:} The IoU values of all matches are summed and then divided by the total number of ground truth boxes plus any additional predicted boxes, which introduces a penalty for extra predictions.

4. \textbf{Dataset-Level IoU Calculation:} The overall IoU for the dataset is computed as the average IoU of all individual images.

\begin{table}[H]
    \centering
    \begin{tabular}{|c|c|c|c|c|c|c|}
        \hline
        & \multicolumn{2}{c|}{Yolo v7} & \multicolumn{2}{c|}{Yolo v8} & \multicolumn{2}{c|}{Faster R-CNN} \\
        \cline{2-7}
        & Base & \textbf{Ours} & Base & \textbf{Ours} & Base & \textbf{Ours} \\
        \hline
        Average IOU & 0.183 & \textbf{0.301} & 0.218 & \textbf{0.345} & 0.313 & \textbf{0.362} \\
        \hline
        mAP@50-95 & 0.0569 & \textbf{0.102} & 0.142 & \textbf{0.312} & 0.135 & \textbf{0.191} \\
        \hline
        mAP@50 & 0.122 & \textbf{0.191} & 0.238 & \textbf{0.504} & 0.251 & \textbf{0.336} \\
        \hline
        Precision & 0.848 & \textbf{0.74} & 0.697 & \textbf{0.86} & 0.544 & \textbf{0.531} \\
        \hline
        F1 Score & 0.220 & \textbf{0.344} & 0.320 & \textbf{0.576} & 0.554 & \textbf{0.588} \\
        \hline
    \end{tabular}
    \caption{Performance comparison between Yolo v7, Yolo v8, and Faster R-CNN}
    \label{tab:performance_comparison}
\end{table}

It is observed that YOLO v8 performs the best among the three models for our task. We note an approximate average performance increase of 28\% for YOLO v7, 41\% for YOLO v8, and 16\% for Faster-RCNN across all evaluation metrics, as detailed in [Table \ref{tab:performance_comparison}].

\subsection{Results Based on Domain Shift and Real Examples}

To demonstrate the robustness of the proposed pipeline, extensive experiments were conducted using real images with domain shifts. The performance of YOLO v8 fine-tuned with the proposed pipeline on snowy images was compared to YOLO v8 fine-tuned solely on the ground truth traffic light dataset. The evaluation involved 50 images each of snowfall, rainfall, fog, smog, and fire-based smoke conditions sourced from Google searches. 
Results indicated that the model fine-tuned using the proposed pipeline achieved superior performance. This improvement is attributed to data-based augmentation, which shifts the training data distribution, enhancing the model's performance in scenarios where the testing data distribution largely differs from the training data, i.e., in cases involving domain shifts.
Specifically, in scenarios of heavy traffic and snowfall, the baseline model showed significant inaccuracies, such as misclassifying bus lights as traffic lights, which is problematic in real-life applications. In high fog and smog conditions, the baseline model failed to detect all traffic lights within an image . Additionally, in images with substantial smoke and low visibility due to rainfall or fire, models trained using our pipeline outperformed the baseline model due to the extensive augmentation in the test set. Detailed results on a selection of real images are shown in [Figure \ref{fig:informative}].

\section{Limitations and Future Work}
The proposed method has some limitations. It may struggle with very low visibility conditions \cite{ouyang2013modeling} and may not perform well with non-standard traffic signal colors, such as brown instead of black or yellow \cite{zhu2012we}. There is room for improvement in these specific scenarios.

\section{Conclusion}

In this study, a novel pipeline was introduced to enhance the training and fine-tuning of object detection models under adverse weather conditions, with a particular focus on traffic light detection in snowy environments. Using the Bosch Small Traffic Light Dataset, artificial snow was applied to the images, and models such as YOLO v7, YOLO v8, and Faster RCNN within the Detectron 2 framework were fine-tuned. Experimental results show that models fine-tuned with our proposed pipeline outperform those trained on traditional datasets in terms of IoU, mAP, and F1 metrics. This method significantly improves model performance, especially in cases of domain shift, contributing to the development of more robust models for critical applications like autonomous driving, thereby enhancing their reliability and safety in challenging weather conditions.

\bibliographystyle{ACM-Reference-Format}
\bibliography{software}




\end{document}